\pdfoutput=1
\documentclass{article}

\usepackage{microtype}
\usepackage{graphicx}
\usepackage{subfigure}
\usepackage{booktabs} 
\usepackage{makecell}
\usepackage{stfloats}

\usepackage{hyperref}

\newcommand{\exper}[1]{\textsc{#1}}

\newcommand{\prefix}{\exper{Prefix}}

\usepackage{amsmath,amsfonts,bm}

\def\eqref#1{equation~\ref{#1}}

\def\1{\bm{1}}

\DeclareMathAlphabet{\mathsfit}{\encodingdefault}{\sfdefault}{m}{sl}
\SetMathAlphabet{\mathsfit}{bold}{\encodingdefault}{\sfdefault}{bx}{n}

\def\sR{{\mathbb{R}}}

\usepackage[accepted]{icml2023}

\usepackage{amsmath}
\usepackage{amssymb}
\usepackage{mathtools}
\usepackage{amsthm}
\usepackage{multirow} 

\usepackage[capitalize,noabbrev]{cleveref}

\theoremstyle{plain}

\theoremstyle{definition}

\theoremstyle{remark}

\usepackage[textsize=tiny]{todonotes}
\usepackage{colortbl}

\definecolor{super-good}{HTML}{ff6700}
\definecolor{very-good}{HTML}{ffa861}
\definecolor{good}{HTML}{ffb38a}
\definecolor{slight-good}{HTML}{ffd2ad}
\definecolor{bit-good}{HTML}{f7efe8}

\definecolor{super-bad}{HTML}{02a9f7}
\definecolor{very-bad}{HTML}{63aee2}
\definecolor{bad}{HTML}{89d6fb}
\definecolor{slight-bad}{HTML}{a5d0ee}
\definecolor{bit-bad}{HTML}{d4f0fc}

\definecolor{textorange}{HTML}{fe8216}
\definecolor{textblue}{HTML}{2178b4}
\icmltitlerunning{PrefixMol: Target- and Chemistry-aware Molecule Design via Prefix Embedding}

\begin{document}

\twocolumn[
    \icmltitle{PrefixMol: Target- and Chemistry-aware Molecule Design via Prefix Embedding}



    \icmlsetsymbol{equal}{*}
    \begin{icmlauthorlist}

        \icmlauthor{Zhangyang Gao}{equal,westlake,zhejiang}
        \icmlauthor{Yuqi Hu}{equal,westlake,sztu}
        \icmlauthor{Cheng Tan}{westlake,zhejiang}
        \icmlauthor{Stan Z. Li}{westlake}
    \end{icmlauthorlist}

    \icmlaffiliation{westlake}{AI Research and Innovation Lab, Westlake University}
    \icmlaffiliation{zhejiang}{Zhejiang University}
    \icmlaffiliation{sztu}{BDI, Shenzhen Technology University, Shenzhen, China}

    \icmlcorrespondingauthor{Stan Z. Li}{Stan.ZQ.Li@westlake.edu.cn}

    \icmlkeywords{Machine Learning, ICML}

    \vskip 0.1in
]



\printAffiliationsAndNotice{\icmlEqualContribution} 

\begin{abstract}
Is there a unified model for generating molecules considering different conditions, such as binding pockets and chemical properties? Although target-aware generative models have made significant advances in drug design,  they do not consider chemistry conditions and cannot guarantee the desired chemical properties. Unfortunately, merging the target-aware and chemical-aware models into a unified model to meet customized requirements may lead to the problem of negative transfer. Inspired by the success of multi-task learning in the NLP area, we use prefix embeddings to provide a novel generative model that considers both the targeted pocket's circumstances and a variety of chemical properties. All conditional information is represented as learnable features, which the generative model subsequently employs as a contextual prompt. Experiments show that our model exhibits good controllability in both single and multi-conditional molecular generation. The controllability enables us to outperform previous structure-based drug design methods. More interestingly, we open up the attention mechanism and reveal coupling relationships between conditions, providing guidance for multi-conditional molecule generation.
\end{abstract}

\vspace{-3mm}
\section{Introduction}
\vspace{-1mm}
\label{sec:intro}
Recently, deep learning methods have shown promising potential for discovering desired drug molecules. Given that the order of drug-like spaces is $10^{60}$ to $10^{100}$ \cite{schneider2005computer}, \textit{de novo} drug discovery is often described as finding a needle in a haystack. In recent years, numerous rule-based algorithms \cite{patel2009knowledge} to explore the chemical space have been presented; however, the computational overhead and results have been far from ideal. Inspired by the success of image, audio, and text generation models \cite{kiros2014multimodal,baltruvsaitis2018multimodal,xu2022multimodal,gao2022simvp,gao2022pifold, tan2023generative, tan2022temporal,tan2022rfold, gao2023diffsds}, researchers have recently turned to deep learning models to generate the desired molecules directly, eliminating the need to search the vast drug-like space \cite{peng2022pocket2mol,liu2022generating}. The critical issue in this type of research is determining how to control the behavior of the model to generate molecules with desired features.

Target-aware generative models have generated considerable attention in AI-assisted drug discovery. As pharmaceutical molecules are only effective when they bind to target proteins, creating molecules with a high affinity to the target is crucial. To meet this requirement, sequence-based\cite{bagal2021molgpt}, graph-based\cite{tan2022target}, and 3D-based\cite{liu2022generating,peng2022pocket2mol,ragoza2022chemsci,luo20223d} generative models that consider protein-ligand interactions are proposed. As these methods can manufacture targeted therapeutic molecules, they are more likely to be applicable in drug discovery. However, they impose no constraints on the chemistry of the generated molecules and are, therefore, unable to control their chemical properties.

Considering the targeted protein in conjunction with multiple chemical properties remains explored. Few models can produce compounds that target specific proteins and simultaneously regulate their chemical properties. The difficulty comes from several aspects: First, there needs to be more high-quality datasets that contain both target-protein affinity and molecular chemical characteristics. Molecules with missing property values further complicate the modeling process due to the absence of labels. Second, it is far more challenging to use a unified model to meet the customized requirements than to generate molecules based on a single condition, as the pocket-aware generative models do. Treating the modelling of each condition as a separate task, the unified model is a mixture of numerous multi-task models, which may suffer from the problem of negative transfer \cite{crawshaw2020multi}: joint training of tasks hurts learning instead of helping it. The challenge is to \textit{develop an effective unified model considering multiple conditions (tasks) such as binding pockets and chemical properties.}

To address the problems above, we extend the CrossDocked data set containing protein-ligand pairs with molecular properties and develop a prefix-conditional model to unify multi-conditional generation. Inspired by the success of multi-task learning in the NLP area \cite{he2021towards,pilault2020conditionally,liu2022few,wu2020understanding,wortsman2022model,li2021prefix,hu2021lora,houlsby2019parameter}, we suggest prepending learnable conditional feature vectors to the query and key of the attention module, resulting in the PrefixMol method. The prefix embedding is always on the left side, serving as a task-related contextual prompt to affect the predicted outcomes on its right. These prefix embeddings are  learned by auxiliary neural networks, considering the 3D pocket and chemical properties.  Experiments show that PrefixMol demonstrates good controllability in both single- and multi-conditional settings. Moreover, the controllability enables us to outperform previous structure-based drug design methods. Last but not least, we open up the attention mechanism and reveal the coupling relationships between conditions, providing guidance for multi-conditional molecule generation.

\vspace{-3mm}
\section{Related Work}

\label{sec:related}

\vspace{-1mm}
\textbf{Problem Definition.} 
Denote $\mathbf{x} \in \mathbb{R}^l$ is the SMILES molecular representation of length $l$, with $n_c$ chemical properties $\mathbf{c} = \{ c_1, c_2, \cdots, c_{n_c} \}$. In our setting, $n_c=6$ and $\{ c_1, c_2, c_3, c_4, c_5, c_6 \} := \{ \text{Pocket}, \text{VINA}, \text{QED}, \text{LogP}, \text{SA}, \text{Lipinski} \}$. Considering the user has a desired ranges of properties $\mathbf{c}^* = \{ c_1^*, c_2^*, \cdots, c_{n_c}^* \}$, controllable molecule generation aims to learn a data generator $g_{\theta}: \mathbf{z}\mapsto \mathbf{x}$, to satisfy the user desires, such that $\mathbf{c} \in \mathbf{c}^*$. Target-aware generative model maximizes $p(\mathbf{x}| c_1)$ through the model $g_{\theta_1}(\textbf{x}, c_1)$ with learnable parameter $\theta_1$, while we aim to merge $\{ g_{\theta_i}(\textbf{x}|c_i) \}_{i=1}^N$ as a unified model $g_{\theta}(\textbf{x}, \mathbf{c})$.

\textbf{Target-aware Molecular Generation.} 
Recently, various molecular generation methods have attracted extensive attention \citep{gebauer2019symmetry, simm2020generative, simm2020reinforcement, shi2021learning, xu2021end, luo2021predicting, xu2020learning, ganea2021geomol, xu2022geodiff, hoogeboom2022equivariant, jing2022torsional, zhu2022direct, RWtoMGonParagraph1.0, nesterov20203dmolnet,gebauer2022inverse,wu2022diffusion,huang2022mdm,huang20223dlinker,wang2022generative,gao2020lookhops, gao2021git, gao2022cosp,gao2022semiretro,xia2021towards,gao2022alphadesign,tan2022generative,tan2022hyperspherical, xia10mole}. However, few methods consider protein-ligand interactions to generate molecules that bind to specific protein targets \citep{imrie2020deep, luo20223d,ragoza2022chemsci,peng2022pocket2mol,liu2022generating}. In Table \ref{tab: mol_generate_models} (Appendix), we divide target-aware molecular generation models into two types: graph-based and 3D structure-based. Graph-based methods generate molecular graphs given the protein sequence information. For example, SiamFlow \citep{tan2022target} develops a flow model to generate molecular graphs given the targeted protein sequence. To better consider the spatial information, such as spatial isomerism and non-bonded interaction, more 3D structure-based methods have been proposed \citep{imrie2020deep, luo20223d, ragoza2022chemsci, peng2022pocket2mol, liu2022generating}. Among them, Pocket2Mol \citep{peng2022pocket2mol} and GraphBP \citep{liu2022generating} are  representative models to autoregressively generate the atoms.

\begin{figure*}[t]
  \centering
  \includegraphics[width=6.8in]{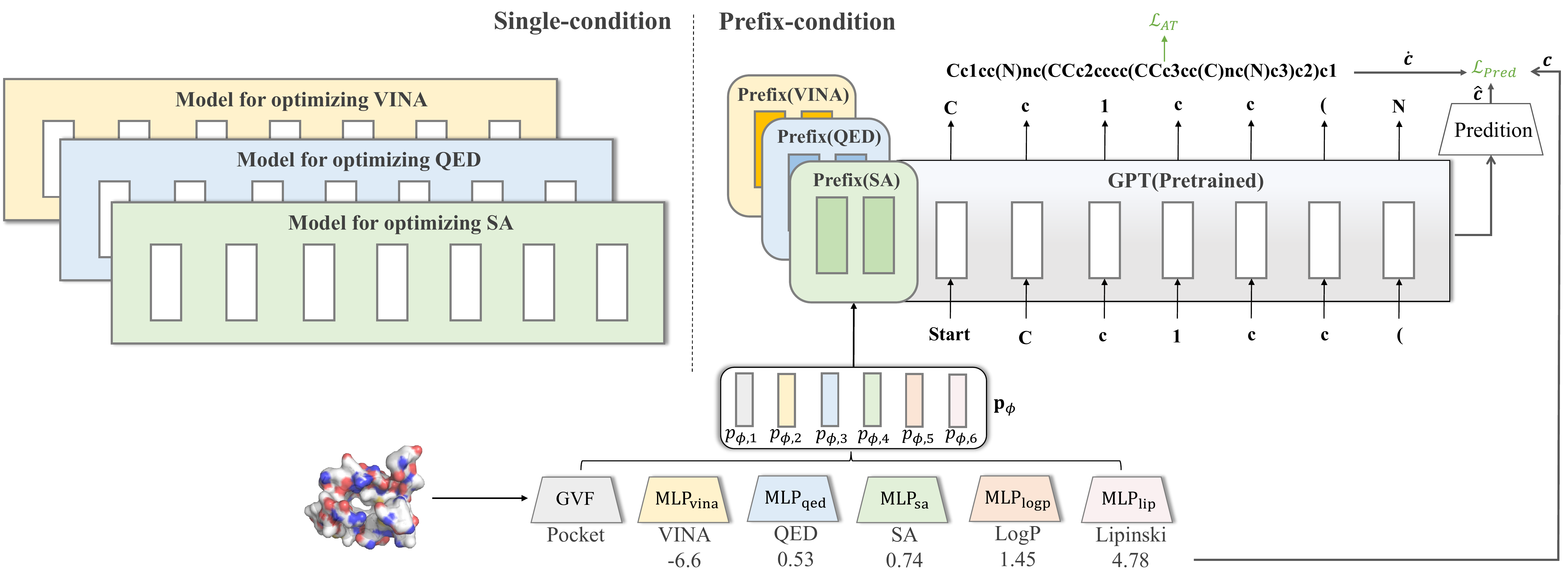}
  \vspace{-6mm}
  \caption{The overall framework. Multiple conditions are embedded as learnable features, including the 3D pocket, Vina Score, QED, SA, LogP, and Lipinski. We append these conditional embeddings in the left side of the sequence, serving as a contextual prompt for a molecular generation. We call this method PrefixMol which allows customized models for generating molecules with single or multiple desired properties by modifying the prefix features. The auto-regressive loss and triplet property prediction loss train PrefixMol.}
  \label{fig:framework}
\end{figure*}

\textbf{Controllable Molecule Generation.}
Although compounds are selected based on their projected bioactivities, their absorption, distribution, metabolism, excretion, and toxicity (ADMET) properties are frequently difficult to predict and adjust, causing bottlenecks in downstream investigations and applications. It would be more productive if candidate molecules with adequate chemical properties were developed at the outset of the molecule design process. Most recent
research \cite{RWtoMOptParagraph1.0,RWtoMOptParagraph1.2,RWtoMOptParagraph1.4,RWtoMOptParagraph1.5,lim2018molecular,shin2021controlled,das2021accelerated,wang2022retrieval} proposes to synthesize molecules in a controlled manner using generative models, which we summarize four different generation approaches shown in Table \ref{tab: mol_controllable_models} (Appendix). On the one hand, reinforcement-learning (RL) and supervised-learning (SL) approaches necessitate extensive task-specific fine-tuning. Optimization-based techniques, on the other hand, train latent-space property predictors to uncover latent information relating to the target molecules. However, in real-world circumstances, we only have a limited amount of active molecules accessible for training. To address these problems, RetMol \cite{wang2022retrieval} proposes a retrieval-based framework for controllable molecule generation. \cite{shin2021controlled} develops CMG extending the self-attention technique Transformer to a molecular sequence by incorporating molecule properties and additional regularization network.

\vspace{-3mm}
\section{Method}
\label{sec:method}

\subsection{Overall Framework}
We propose PrefixMol, inserting learnable conditional feature vectors into the attention module to unify multi-conditional molecule generative models to support the modeling of customized requirements. We illustrate the overall framework in Figure \ref{fig:framework}. Compared to previous conditional generative models, our innovations include the following:

\vspace{-2mm}
\begin{enumerate}
  \vspace{-2mm}
  \item Extending the CrossDocked data set with molecular properties for multi-conditional generation.
        \vspace{-2mm}
  \item Proposing PrefixMol to support multi-condition modeling for meeting customized requirements.
        \vspace{-2mm}
  \item Providing insight into how the conditions control the model behavior and correlate with each other.
        \vspace{-2mm}
  \item Conducting systematic experiments to evaluate the proposed method.
        \vspace{-2mm}
\end{enumerate}

\subsection{Prefix-conditional GPT}
\textbf{GPT Transformer.}
We adopt the GPT model \cite{brown2020language} to generate the molecular SMILES string, where the critical component is the transformer layer (Attention+FFN). The output of the $k$-th multi-head attention (MHA) layer presents as follows.

\vspace{-4mm}
\begin{equation}
  \begin{aligned}
    \begin{cases}
      \text{Attn}(\mathbf{q},\mathbf{k},\mathbf{v}) = \text{softmax}(\frac{\mathbf{q}\mathbf{k}^T}{\sqrt{d_k}}) \mathbf{v} \\
      \mathrm{MHA}(\mathbf{x}, \mathbf{c}) =  \mathrm{Cat(head_1, \cdots, head_h)} \mathbf{W}_o                            \\
      \mathrm{head_i} = \mathrm{Attn}(\mathbf{x} \mathbf{W}_q^{(i)}, \mathbf{c} \mathbf{W}_k^{(i)}, \mathbf{c} \mathbf{W}_v^{(i)}),
      \label{eq: multi-head: attn}
    \end{cases}
  \end{aligned}
\end{equation}
\vspace{-4mm}

where $\mathbf{x} \in \mathbb{R}^{l\times d}$ is a sequence of $l$ vectors over which we would like to perform attention. The $i$-th head is parameterized by $\mathbf{W}_q^{(i)}, \mathbf{W}_k^{(i)}, W_v^{(i)} \in \mathbb{R}^{d\times d_h}$ to project inputs to queries, keys, and values.  $W_o \in \sR^{d_h\times d}$ projects features into the model dimension. Later on, the attention output $\textbf{z} = \text{Attn}(\mathbf{q},\mathbf{k},\mathbf{v})$ are fed into a fully connected feed-forward network (FFN):

\vspace{-4mm}
\begin{equation}
  \begin{aligned}
    \mathrm{FFN}(\mathbf{z}) = \mathrm{ReLU}(\mathbf{z} \mathbf{W}_1 + \mathbf{b}_1) \mathbf{W}_2 + \mathbf{b}_2, \\
  \end{aligned}
\end{equation}

\vspace{-2mm}

where $\mathbf{W}_1 \in \mathbb{R}^{d\times d_m}$, $\mathbf{W}_2 \in \sR^{d_m\times d}, d_m=4d$. Finally, a residual connection is used, followed by layer normalization.

\textbf{Prefix Conditional Embeddings.}
Upon the original sequence embedding $\textbf{x} \in \mathbb{R}^{l,d}$ of length $l$, we suggest prepending conditional features on the left, resulting in extended input $\mathbf{x}' = [\prefix; \mathbf{x}]$, as shown in Figure \ref{fig:framework}. We use an additional learnable matrix $\mathbf{p}_{\phi} \in \mathbb{R}^{n_c,d}$ to store the learnable prefix parameters. For simplicity, we write the $i$-th prefix feature vector as $p_{\phi,i}$. The output features of the language model will be:
\begin{equation}
  \begin{aligned}
    h_{i} =
    \begin{cases}
      p_{\phi,i},                     & \text{if } i < n_c \\
      \text{LM}_\theta(x'_i, h_{<i}), & \text{otherwise.}
    \end{cases}
  \end{aligned}
\end{equation}
where $x'_i$ is the $i$-th element of the extended input,  $h_i$ is the $i$-th output feature, $\phi$ and $\theta$ are learnable parameters. The prefix condition features are always on the left context and therefore affect any predictions on its right. More importantly, this approach decouples task-specific ($\phi$) and generic ($\theta$) knowledge, allowing the user to apply different tasks by modifying the conditional vector.

\textbf{Condition Controlling \& Correlation.}
By analyzing the extended attention layer, we can determine: (1) how the conditions affect the behavior of the model and (2) how the conditions are interrelated. As to the first question, we provide the formula derivation procedure and variable declarations in the Appendix, reformulating the attention mechanism for the sequence embedding $\textbf{x}$ as follows:

\vspace{-4mm}
\begin{equation}
  \begin{aligned}
    \label{eq:prefix-adapter}
    \small
    \begin{split}
      & head = (1 - \lambda(\mathbf{x})) \underbrace{ \text{Attn}(\mathbf{x}\mathbf{W}_q, \mathbf{c}\mathbf{W}_{k}, \mathbf{c}\mathbf{W}_v) }_{\text{self attention}}\\ & \quad + \lambda(\mathbf{x}) \underbrace{ \text{Attn}(\mathbf{x}\mathbf{W}_q, \mathbf{p}_{\phi} \mathbf{W}_{k}, \mathbf{p}_{\phi} \mathbf{W}_{v}) }_{\text{prefix attention }}
    \end{split}
  \end{aligned}
\end{equation}
\vspace{-3mm}

From Equation \ref{eq:prefix-adapter}, we know that the prefix conditions control the model behavior by modifying the original attention weights. Thus, the activation map derived from $\text{softmax}(\mathbf{x}\mathbf{W}_q \mathbf{W}_{k}^\top\mathbf{p}_{\phi}^\top)$ could be used for analyzing how the conditions control the model behavior. Similarly, we re-formulate the attention computation of the prefix features $\mathbf{p}_{\phi}$, detailed in Appendix \ref{ap:prefix_adp} and written as:

\vspace{-4mm}
\begin{equation}
  \label{eq:correlation}
  \begin{aligned}
    \small
    \begin{split}
      & head = \underbrace{\text{Attn}(\mathbf{p}_{\phi}\mathbf{W}_q, \mathbf{p}_{\phi}\mathbf{W}_k, \mathbf{p}_{\phi}\mathbf{W}_v)}_{\text{prefix correlation}}
    \end{split}
  \end{aligned}
\end{equation}
\vspace{-3mm}

where the $\text{Attn}(\mathbf{p}_{\phi} \mathbf{W}_k, \mathbf{c}\mathbf{W}_{k}, \mathbf{c}\mathbf{W}_v)$ term equal to zero, due to the causal mask applied on the model. The remaining term, i.e., $\text{Attn}(\mathbf{p}_{\phi}\mathbf{W}_q, \mathbf{p}_{\phi}\mathbf{W}_k, \mathbf{p}_{\phi}\mathbf{W}_v)$, reveals how  prefix features correlate with each other.

\textbf{Auto-regressive Loss.}
Similar to GPT-3 \cite{brown2020language}, we use the auto-regressive loss to train the generative model:

\vspace{-5mm}
\begin{equation}
  \begin{aligned}
    \mathcal{L}_{AT} & = -\min_{\phi, \theta} ~\log p_{\phi, \theta}(\mathbf{x}_{1:t} \mid \mathbf{x}_{0:t-1}, \mathbf{p}_{\phi}) \\
                     & \quad = -\sum_{1< i\leq t}  \log p_{\phi, \theta}(x_{i} \mid \textbf{x}_{<i}, \mathbf{p}_{\phi})
    \label{eqn:loss}
  \end{aligned}
\end{equation}
\vspace{-3mm}

The difference is that we introduce prefix conditions during the generative process.

\vspace{-2mm}
\paragraph{Property Prediction Loss.} In addition to the auto-regressive loss, we impose triplet predictive loss upon the model for generating molecules with desired properties:

\vspace{-5mm}
\begin{equation}
  \begin{aligned}
    \mathcal{L}_{Pred} & = \max(( \hat{\textbf{c}} - \textbf{c} )^2 - (\hat{\textbf{c}} - \dot{\textbf{c}})^2, 0)
    \label{eqn:pred_prop}
  \end{aligned}
\end{equation}
\vspace{-6mm}

Where $\textbf{c}$ is the input properties serving as conditions, $\hat{\textbf{c}}$ is the properties calculated by an MLP prediction head, and $\dot{\textbf{c}}$ is the properties of the generated molecule. The triplet loss requires the model to generate molecules whose properties are consistent with the input conditions. Since $\dot{\textbf{c}}$ is computed by RDKit according to the generated SMILES and is non-differentiable, we propagate the gradient with the help of $\hat{\textbf{c}}$.

\subsection{Conditional Embeddings}

\textbf{3D Pocket Embedding.}
We propose GVF (Geometric Vector Transformer), a variant of GVP (Geometric Vector Perceptrons) GNN \cite{jing2020learning},  to extract 3D pocket features. Consider a pocket that has $n_v$ atoms, we represent it as a 3D graph $\mathcal{G}(\mathcal{V}, \vec{\mathcal{V}}, \mathcal{E}, \vec{\mathcal{E}})$ , consisting of node features ($\mathcal{V} \in \mathbb{R}^{n_v, d_f}$, $\vec{\mathcal{V}} \in \mathbb{R}^{n_v, 3}$) and edge features ($\mathcal{E} \in \mathbb{R}^{n_v, d_e}$, $\vec{\mathcal{E}} \in \mathbb{R}^{n_v, 3}$). Note that $\mathcal{V}$ and $\mathcal{E}$ are invariant features with dimension $d_f$ and $d_e$, respectively, and $\vec{\mathcal{V}}$, $\vec{\mathcal{E}}$ are equivariant geometric features. Previous works \cite{jing2020learning, peng2022pocket2mol} have shown that considering both scalar ($\mathcal{V}, \mathcal{E}$) and vector features ($\vec{\mathcal{V}}, \vec{\mathcal{E}}$) helps the model to learn expressive 3D representations. However, all these methods only consider local interactions through graph message passing while ignoring global contextual interactions, which may limit the expressive power of the model. As a remedy, we introduce a new GVF layer considering both local and global geometric interactions by adding a global attention module upon the GNN layer. The GVF layer is formulated as follows:

\vspace{-3mm}
\begin{equation}
  \begin{aligned}
    \label{eq:GVFormer}
    \begin{cases}
      (\mathcal{V}, \vec{\mathcal{V}}) & = \text{GNN}(\mathcal{V}, \vec{\mathcal{V}}, \mathcal{E}, \vec{\mathcal{E}})                           \\
      A                                & = \text{Softmax}(\frac{\vec{\mathcal{V}}'^T \mathbf{W}_q^T \mathbf{W}_k \vec{\mathcal{V}}'}{\sqrt{d}}) \\
      ({\mathcal{V}}', {\mathcal{E}})  & \leftarrow (\text{FFN}_{v}(A{\mathcal{V}}'), \text{FFN}_{e}{(A{\mathcal{V}}'||\mathcal{E})})           \\
    \end{cases}
  \end{aligned}
\end{equation}

Where $\mathbf{W}_q$ and $\mathbf{W}_k$ are learnable global attention parameters, $\text{FFN}_{v}$ and $\text{FFN}_{e}$ are feed-forward MLPs that transform node and edge features. We use the same GNN architecture as Pocket2Mol \cite{peng2022pocket2mol} for considering local interactions. We randomly choose an anchor node $v_i$ that is within 5 $\dot{A}$ to the bounded ligand molecule and computing the pocket embedding $p_{1}$ through position-aware attention:

\vspace{-3mm}
\begin{equation}
  \begin{aligned}
    h_1 = \sum_{j=1}^{n_v} \text{MLP}_{att}(\text{rbf}(d_{ij}) ||  v_i || v_j)) v_i
  \end{aligned}
\end{equation}

where $\text{MLP}_{att}$ is an MLP used for computing attention weights, $\text{rbf}(\cdot)$ is a radial basis function, $||$ indicates the concatenate operation.

\textbf{Property Embedding.}
In addition to the 3D pocket condition, we also consider multiple chemistry properties as conditions, including Vina ($c_2$), QED ($c_3$), SA ($c_4$), LogP ($c_5$) and Lipinski ($c_6$); see the experiment section for more details about these properties. We use separate MLPs to embed each property, formulated as follows:
\begin{equation}
  \begin{aligned} \label{eq:GVFormer}
    \begin{cases}
      p_{2} & = \text{MLP}_{vina}(c_2) \\
      p_{3} & = \text{MLP}_{qed}(c_3)  \\
      p_{4} & = \text{MLP}_{sa}(c_4)   \\
      p_{5} & = \text{MLP}_{logp}(c_5) \\
      p_{6} & = \text{MLP}_{lip}(c_6)  \\
    \end{cases}
  \end{aligned}
\end{equation}
\section{Experiments}
\label{sec:experiments}

\subsection{Experimental Settings}
In this section, we conduct extensive experiments to evaluate the proposed method. Specifically, we would like to answer the following questions:

\textbf{Q1: Comparision.} How does PrefixMol perform compared to previous structure-based drug design methods without conditions?

\textbf{Q2: Controllability.} Could PrefixMol outperform baselines with controllable  conditions? How well does it work in single and multi-conditional settings?

\textbf{Q3: Condition relations.} Are there coupling relationships between control conditions? 

\vspace{-2mm}
\subsection{Basic Settings}
\textbf{Data Set.} 
We use the CrossDocked data set \cite{CrossDocked} with 22.5 million protein-ligand structures to evaluate the proposed method, in which we add chemical properties for each ligand. We follow the same data splitting and evaluation protocols as \cite{peng2022pocket2mol} and \cite{masuda2020generating}.

\vspace{-3mm}
\paragraph{Metrics.} 
To measure the quality of the generated drug candidates, we adopt the following widely known metrics, including \textbf{VINA}, \textbf{QED}, \textbf{SA}, \textbf{LogP}, and \textbf{Lipinski}. We provide a detailed explanation of these metrics in the appendix. In addition, three additional metrics are included for assessing each binding site's generational quality and diversity:  (1) \textbf{High Affinity} is the proportion of pockets whose generated molecules have higher or equal affinities than those in the test set. (2) \textbf{Diversity} \cite{jin2020composing} quantifies the diversity of compounds synthesized for a binding site. It is computed by average pairwise Tanimoto similarity \cite{bajusz2015tanimoto, tanimoto1958elementary} over Morgan fingerprints for all produced molecules of a target. (3) \textbf{Sim.Train} indicates the most related molecules in training set for Tanimoto similarity.
In our work, VINA is calculated by QVina  \cite{trott2010autodock, alhossary2015qvnia} to compute the binding affinity. Before putting the molecules into the calculation of  Vina score, we use universal force fields (\textit{UFF} \cite{rappe1992uff}) to refine the produced structures according to \cite{masuda2020ligan}. Other chemical properties can be calculated by RDKit \cite{rdkit}.

\begin{table}[h]
    \caption{Comparing the properties of the molecules in the test set to those generated by algorithms. Here we present the unconditional version of PrefixMol. The \textbf{best} and \underline{suboptimal} results are labeled with bold and underlined.}
    \label{tab:gen_metrics}
    \centering
    \resizebox{1.0 \columnwidth}{!}{
    \begin{tabular}{c|ccccc}
        \toprule
        Metrics        & \makecell[c]{Test                       \\Set} & CVAE & AR &  \makecell[c]{Pocket2\\Mol} & \makecell[c]{PrefixMol\\(unconditional)} \\ \midrule
        \makecell[c]{VINA                                        \\(kcal/mol, ↓)} & \makecell[c]{-7.158\\ $\pm$ 2.10} & \makecell[c]{-6.144\\ $\pm$ 1.57} & \makecell[c]{-6.215\\ $\pm$ 1.54} & {\makecell[c]{\textbf{-7.288} \\ $\pm$ 2.53}} &  {\makecell[c]{\underline{-6.532} \\ $\pm$ 1.76}}\\ \hline
        QED (↑)        & \makecell[c]{0.484                      \\$\pm$0.21} & \makecell[c]{0.369\\$\pm$0.22} & \makecell[c]{0.502\\$\pm$0.17} & {\makecell[c]{\textbf{0.563}\\$\pm$0.16}}  &  {\makecell[c]{\underline{0.551} \\ $\pm$ 0.18}}\\ \hline
        SA (↑)         & \makecell[c]{0.732                      \\$\pm$0.14} & \makecell[c]{0.590\\$\pm$0.15} & \makecell[c]{0.675\\$\pm$0.14} & {\makecell[c]{\textbf{0.765}\\$\pm$0.13}} &  {\makecell[c]{\underline{0.750} \\ $\pm$ 0.09}}\\ \hline
        LogP           & \makecell[c]{0.947                      \\$\pm$2.65} & \makecell[c]{-0.140\\$\pm$2.73} & \makecell[c]{0.257\\$\pm$2.01} & {\makecell[c]{\textbf{1.586}\\$\pm$1.82}} &  {\makecell[c]{\underline{1.415} \\ $\pm$ 2.11}}\\ \hline
        Lipinski (↑)   & \makecell[c]{4.367                      \\$\pm$1.14} & \makecell[c]{4.027\\$\pm$1.38} & \makecell[c]{\underline{4.787}\\$\pm$0.50} & {\makecell[c]{\textbf{4.902}\\$\pm$0.42}} &  {\makecell[c]{4.710 \\ $\pm$ 0.63}}\\ \hline
        \makecell[c]{High Affinity                               \\(\%, ↑)} & - & \makecell[c]{0.238} & \makecell[c]{0.267} & \makecell[c]{\textbf{0.542}}   &  \makecell[c]{\underline{0.432}}\\ \hline
        Diversity (↑)  & -                  & \makecell[c]{0.654 \\$\pm$0.12} & {\makecell[c]{0.742\\$\pm$0.09}} & \makecell[c]{\underline{0.688}\\$\pm$0.14} & \makecell[c]{\textbf{0.856}\\$\pm$0.17}\\ \hline
        Sim. Train (↓) & -                  & \makecell[c]{0.460 \\$\pm$0.18} & \makecell[c]{0.409\\$\pm$0.19} & {\makecell[c]{\underline{0.376}\\$\pm$0.22}} &
        {\makecell[c]{\textbf{0.239}                             \\$\pm$0.07}}\\
        \bottomrule
    \end{tabular}

}

    \vspace{-1 em}
\end{table}

\begin{table*}[h]
    \caption{\textbf{PrefixMol Single-property Control.}
        Our method \textit{PrefixMol} is evaluated with the conditions for VINA, QED, SA, LogP, and Lipinski.
        The method column \textit{Pocket2Mol} provided in the table compares the controlling effect, and the outcomes surpassing the comparative method \textit{Pocket2Mol} are bolded.
        Colors indicate the performance \textcolor{textblue}{inferior} (lower baseline) or \textcolor{textorange}{superior} (lower baseline) and (+ or -) represent the relative amounts of the baseline.
    }
    \label{tab:single_prop}
    \begin{center}
            \begin{sc}
                \setlength{\tabcolsep}{4.5pt}
                \resizebox{2.0 \columnwidth}{!}{
    \begin{tabular}{lc|ccccc||ccccc||c}
        \toprule
                      & \multicolumn{1}{c}{Baseline} & \multicolumn{5}{c}{Negative}         & \multicolumn{5}{c}{Positive}          & \multicolumn{1}{c}{Method}                                                                                                                                                                                                                                                                                                                                                                                \\

        Metrics       & \multicolumn{1}{c}{0}        & -5                                   & -4                                    & -3                                    & -2                                    & -1                                    & 1                                            & 2                                              & 3                                              & 4                                            & 5                                             & \textit{Pocket2Mol}               \\  \midrule

        VINA (↓)      & -6.532                       & \cellcolor{slight-bad}-6.518(+0.014) & \cellcolor{slight-bad}-6.515(+0.017) & \cellcolor{bit-good}-6.522(+0.01) & \cellcolor{bad}-6.497(+0.035) & \cellcolor{bad}-6.505(+0.027) & \cellcolor{bad}-6.502(+0.03)              & \cellcolor{very-bad} -6.442(+0.09)            & \cellcolor{good}-6.541(-0.009)           & \cellcolor{very-good}-6.552(-0.02)         & \cellcolor{super-bad}-6.351(+0.181)                & -7.288                            \\

        QED (↑)       & 0.551                        & \cellcolor{slight-bad}0.473(-0.078)  & \cellcolor{bit-bad}0.520(-0.031)      & \cellcolor{very-bad}0.354(-0.197)     & \cellcolor{super-bad}0.303(-0.248)    & \cellcolor{bad}0.456(-0.095)          & \cellcolor{very-good} \textbf{0.757}(+0.206) & \cellcolor{slight-good} \textbf{0.732}(+0.181) & \cellcolor{super-good} \textbf{0.767}(+0.216)  & \cellcolor{good} \textbf{0.754}(+0.203)      & \cellcolor{super-good} \textbf{0.767}(+0.216) & 0.563                             \\

        SA (↑)        & 0.750                        & --                                   & --                                    & --                                    & --                                    & --                                    & \cellcolor{bit-good} \textbf{0.889} (+0.139) & \cellcolor{slight-good} \textbf{0.911}(+0.161) & \cellcolor{good} \textbf{0.913}(+0.163)        & \cellcolor{very-good} \textbf{0.920}(+0.170) & \cellcolor{super-good} \textbf{0.924}(+0.174) & 0.765                             \\

        LogP          & 1.415                        & \cellcolor{super-bad}-1.062(-2.477)  & \cellcolor{very-bad}-0.804(-2.219)    & \cellcolor{slight-bad}0.486(-0.929)   & \cellcolor{bad}0.056(-1.359)          & \cellcolor{bit-good}0.676(-0.739)     & \cellcolor{bit-good} \textbf{2.191} (+0.776) & \cellcolor{slight-good} \textbf{2.930}(+1.515) & \cellcolor{good} \textbf{3.358}(+1.943)        & \cellcolor{very-good} \textbf{3.395}(+1.980) & \cellcolor{super-good} \textbf{3.631}(+2.216)
                      & 1.586                                                                                                                                                                                                                                                                                                                                                                                                                                                                                                                   \\

        Lipinski (↑)  & 4.710                        & \cellcolor{slight-bad}4.676(-0.034)    & \cellcolor{bit-bad}4.7(-0.01)    & \cellcolor{good}4.86(+0.15)          & \cellcolor{slight-good}4.759(+0.049)         & 4.710(--)  & \cellcolor{bit-good} 4.721(+0.011)         & \cellcolor{bit-good} 4.721(+0.011)        & \cellcolor{bit-bad} 4.7(-0.01)           & \cellcolor{slight-bad} 4.672(-0.038)       & \cellcolor{slight-bad} 4.677(-0.033)
                      & 4.902 \\ \midrule

        VINA (↓,QED)  & -6.498                       & \cellcolor{very-bad}-6.000(-0.498)   & \cellcolor{bit-bad}-6.325(-0.173)     & \cellcolor{slight-good}-6.220(-0.278) & \cellcolor{bad}-6.011(-0.487)         & \cellcolor{super-bad}-5.942(-0.556)   & \cellcolor{bit-good}-6.876(+0.378)           & \cellcolor{super-good} \textbf{-7.733}(+1.235) & \cellcolor{very-good}\textbf{-7.5625}(+1.0645) & \cellcolor{slight-good}-7.150(+0.652)        & \cellcolor{good}-7.256(+0.758)                & -7.288                            \\

        VINA (↓,LogP) & -6.498                       & \cellcolor{very-bad}-6.229(-0.269)   & \cellcolor{super-bad}-6.100(-0.398)   & \cellcolor{bit-bad}-6.490(-0.008)     & \cellcolor{slight-bad}-6.325(-0.173)  & \cellcolor{bad}-6.315(-0.183)         & \cellcolor{bit-good}-6.820(+0.322)           & \cellcolor{very-good} \textbf{-7.377}(+0.879)  & \cellcolor{super-good}\textbf{-7.465}(+0.967)  & \cellcolor{slight-good}-7.210(+0.712)        & \cellcolor{good}-7.250(+0.752)                & -7.288                            \\
        \bottomrule
    \end{tabular}
}
                \setlength{\tabcolsep}{6pt}
            \end{sc}
    \end{center}
    \vspace{-5mm}
\end{table*}

\subsection{Comparison (Q1)}
\textbf{Objective \& Setting.} 
How does PrefixMol perform compared to previous structure-based drug design methods without conditions? We train the unconditional PrefixMol, where conditions do not use for controlling the generation. We compare PrefixMol with recent strong baselines, including CVAE \cite{masuda2020generating}, AR \cite{luo20213d}, and Pocket2Mol \cite{peng2022pocket2mol}.

\vspace{-3mm}
\textbf{Results \& Analysis.}
The mean values and standard deviations of the measures above are presented in Table \ref{tab:gen_metrics}, with the Prefix version lacking conditional inputs. Probably because PrefixMol (unconditional) does not explicitly model the 3D ligand structure and molecular properties, it could only achieve sub-optimal results on VINA, QED, SA, and LogP. Interestingly, the metrics \textit{Sim. Train} and \textit{Diversity} of PrefixMol exceed other computational models. This phenomenon suggests that PrefixMol is not simply memorizing training data, and that it is more capable than baselines at producing novel molecules.

\begin{figure*}[b]
  \centering
  \includegraphics[width=6.2in]{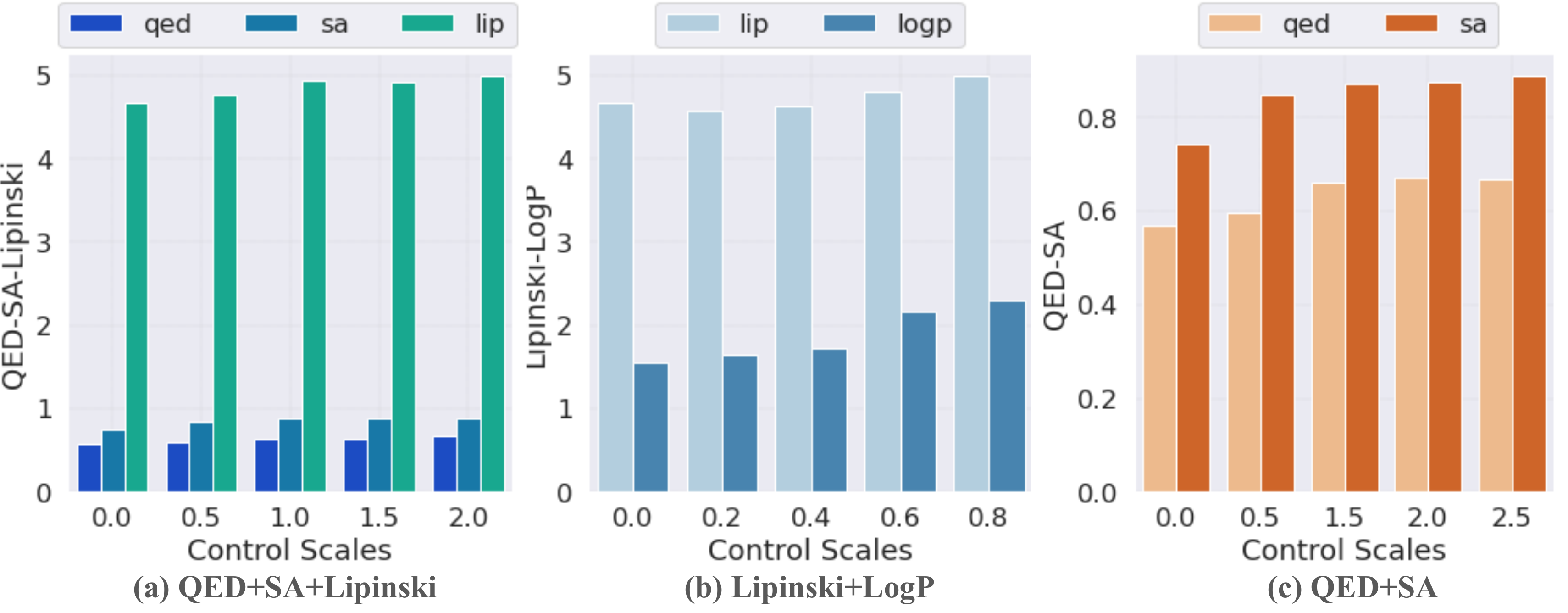}
  \vspace{-3mm}
  \caption{We present condition relations of the metrics (a). Distribution of property of generated molecules conditioned on (b) QED+SA, (c) Lipinski+LogP and (d) QED+SA+Lipinski. The values that the generation is conditioned to are given in the legends of the panels.}
  \vspace{-3mm}
  \label{fig:multi_control}
\end{figure*}

\vspace{-2mm}
\subsection{Controllability (Q2)} 

\vspace{-2mm}
\textbf{Objective \& Setting.}
Numerous biological and chemical processes necessitate that molecules possess specific property values to execute certain functions. Therefore, we aim to analyze PrefixMol's ability to produce molecules with specific properties (controllable generation). We illustrate our model's capabilities of single-property control and multi-properties control to see whether the changes of the conditional inputs (represented as control scales in the tables) substantially impact generating desired molecules. We provide the following example for understanding the control scale: if the ground truth SA is 0.7 and the control scale is 0.05, then the modified SA condition is (0.7 + 0.05).

\textbf{Single-property Control.}
In Table \ref{tab:single_prop}, we depict the effect of controlling a single property, such as VINA, QED, SA, LogP, and Lipinski (presented in mean values). Most cases show that molecular properties are positively correlated with conditional inputs, indicating that the proposed method could effectively control single property. Interestingly, Table \ref{tab:single_prop} further demonstrates that PrefixMol could outperform the SOTA approaches with enhanced control conditions in terms of QED, SA, and LogP. This finding validates the superiority and efficacy of our model's controllability. As QED and LogP are respectively increased in rows VINA(↓, QED) and VINA(↓, LogP), the VINA likewise climbs and surpasses the SOTA baseline, indicating that there are coupling relationships between conditions affecting the control effects.

\begin{table}[htbp]
    \vspace{-3mm}
    \caption{\textbf{PrefixMol Multi-properties Control.}
        Generated molecules conditioned on five metrics verify the controllability.
        Colors indicate the performance of the controlling effect, and the values are shown as the combination of the mean average deviation(MAD) and standard deviation(SD).
    }
    \vspace{3mm}
    \label{tab:multi_prop}
    \centering
    \resizebox{1.0 \columnwidth}{!}{
    \begin{tabular}{l|ccccc}
        \toprule
        \makecell[c]{Control                                                                                                                                                                                     \\Scales} & VINA (↓)                                 & QED (↑)                                 & SA (↑)                                  & LogP                                    & Lipinski (↑)               \\  \midrule
        -4(all) & \cellcolor{bit-good}-4.567($\pm$0.82)    & \cellcolor{bit-good}0.286($\pm$0.11)    & \cellcolor{bit-good}0.557($\pm$0.02)    & \cellcolor{bit-good}-1.494($\pm$0.85)   & \cellcolor{bit-good}3.7    \\
        0(all)  & \cellcolor{slight-good}-6.220($\pm$1.12) & \cellcolor{slight-good}0.547($\pm$0.18) & \cellcolor{slight-good}0.755($\pm$0.07) & \cellcolor{slight-good}0.796($\pm$1.96) & \cellcolor{slight-good}4.8 \\
        +4(all) & \cellcolor{good}-6.333($\pm$0.70)        & \cellcolor{good}0.722($\pm$0.006)       & \cellcolor{good}0.913($\pm$0.04)        & \cellcolor{good}2.433($\pm$1.25)        & \cellcolor{good}5.0        \\ \midrule
         +0.5(QED,SA) & \cellcolor{bit-good}-5.888($\pm$1.46)    & \cellcolor{bit-good}0.596($\pm$0.16)    & \cellcolor{bit-good}0.847($\pm$0.08)    & \cellcolor{bit-good}1.328($\pm$2.06)   & \cellcolor{bit-good}\textbf{4.75}    \\
        +1.5(QED,SA)  & \cellcolor{slight-good}-5.742($\pm$1.60) & \cellcolor{slight-good}0.660($\pm$0.12) & \cellcolor{slight-good}0.872($\pm$0.04) & \cellcolor{slight-good}1.530($\pm$1.10) & \cellcolor{slight-good}\textbf{4.917}\\
        +2.0(QED,SA) & \cellcolor{good}-5.715($\pm$0.65)        & \cellcolor{good}0.671($\pm$0.09)       & \cellcolor{good}0.876($\pm$0.04)        & \cellcolor{good}1.320($\pm$0.78)        & \cellcolor{good}\textbf{5.0}        \\
        +2.5(QED,SA) & \cellcolor{good}-5.589($\pm$0.71)        & \cellcolor{good}0.667($\pm$0.09)       & \cellcolor{good}0.89($\pm$0.04)        & \cellcolor{good}1.511($\pm$0.62)        & \cellcolor{good}\textbf{5.0}        \\
        \bottomrule
    \end{tabular}
}
    \vspace{-5mm}
\end{table}

\textbf{Multi-properties Control.}
We investigate whether PrefixMol is effective at controlling many characteristics and emphasize multi-condition modeling is challenging due to the fact that modeling each condition is a distinct work and multi-task models would be susceptible to negative transfer. Consequently, few approaches could jointly maximize diverse molecular properties. We show in Figure \ref{fig:multi_control} that PrefixMol performs well when jointly controlling two or three properties, then simultaneously change all the input conditions and report results in Table \ref{tab:multi_prop}. The experimental results reveal that all properties fluctuate with control conditions and have consistent positive relationships, indicating that PrefixMol has good controllability in the multi-conditional generation. We also observe that Lipinski is coupled to QED and SA with a saturation value of 5.0 when both QED and SA are at least +2.

\textbf{Visualization.} 
Several examples of generated molecules with higher binding affinities (lower VINA) than the corresponding reference compounds and four case studies are shown in Appendix \ref{ap:molecule_design}. Our generated molecules with more excellent affinity structures differ significantly from reference molecules, suggesting our method can generate novel compounds that bind target proteins rather than just copying or changing reference molecules.

\vspace{-3mm}
\subsection{Condition relations (Q3)}

\vspace{-1mm}
\textbf{Objective \& Setting.} 
As derived in Equation.\ref{eq:correlation}, $\text{Attn}(\mathbf{p}_{\phi}\mathbf{W}_q, \mathbf{p}_{\phi}\mathbf{W}_k, \mathbf{p}_{\phi}\mathbf{W}_v)$ reveals how  prefix features correlate with each other. We write the corresponding attention map as $\textbf{A}(\mathbf{p}_{\phi})=\text{softmax}(\textbf{P}_{\phi}\textbf{W}_q \textbf{W}_k^\top \textbf{P}_{\phi}^\top)$, and add perturbations on input conditions to see how the attention map changes and how they are interrelated.  Recall that $\textbf{A}(\textbf{P}_{\phi})$ is a function of input conditions, which could be rewritten as $\textbf{A}(c_1, c_2, c_3, c_4, c_5, c_6)$, where $c_1$ indicates the protein pocket, $(c_2, c_3, c_4, c_5, c_5, c_6)$ represents other input conditions whose corresponding condition types are $(\text{VINA}, \text{QED}, \text{LogP}, \text{SA}, \text{Lipinski})$. Taking VINA as an example, the partial differentiation of the attention map is 

\vspace{-6mm}
\begin{equation*}
\small
\begin{aligned}
        \frac{\partial{\textbf{A}}}{\partial{c_2}}
        = \frac{\textbf{A}(c_1, c_2+\Delta, c_3, c_4, c_5, c_6) - \textbf{A}(c_1, c_2, c_3, c_4, c_5, c_6)}{\Delta}
\end{aligned}
\end{equation*}
\vspace{-7mm}

where we set $\Delta=1$. We take the absolute values of all partial differentiations and add them together to obtain the relation matrix $\textbf{R}$:

\vspace{-6mm}
\begin{equation}
\begin{aligned}
    \textbf{R} = \sum_{i=2}^{6} |\frac{\partial{\textbf{A}}}{\partial{c_i}}|
\end{aligned}
\end{equation}
\vspace{-6mm}

As the causal mask is applied to the attention model for auto-regressive generation, the relation matrix $\textbf{R}$ computed from attention maps is triangular.

\begin{figure}[h]
    \centering
    \includegraphics[width=2.5in]{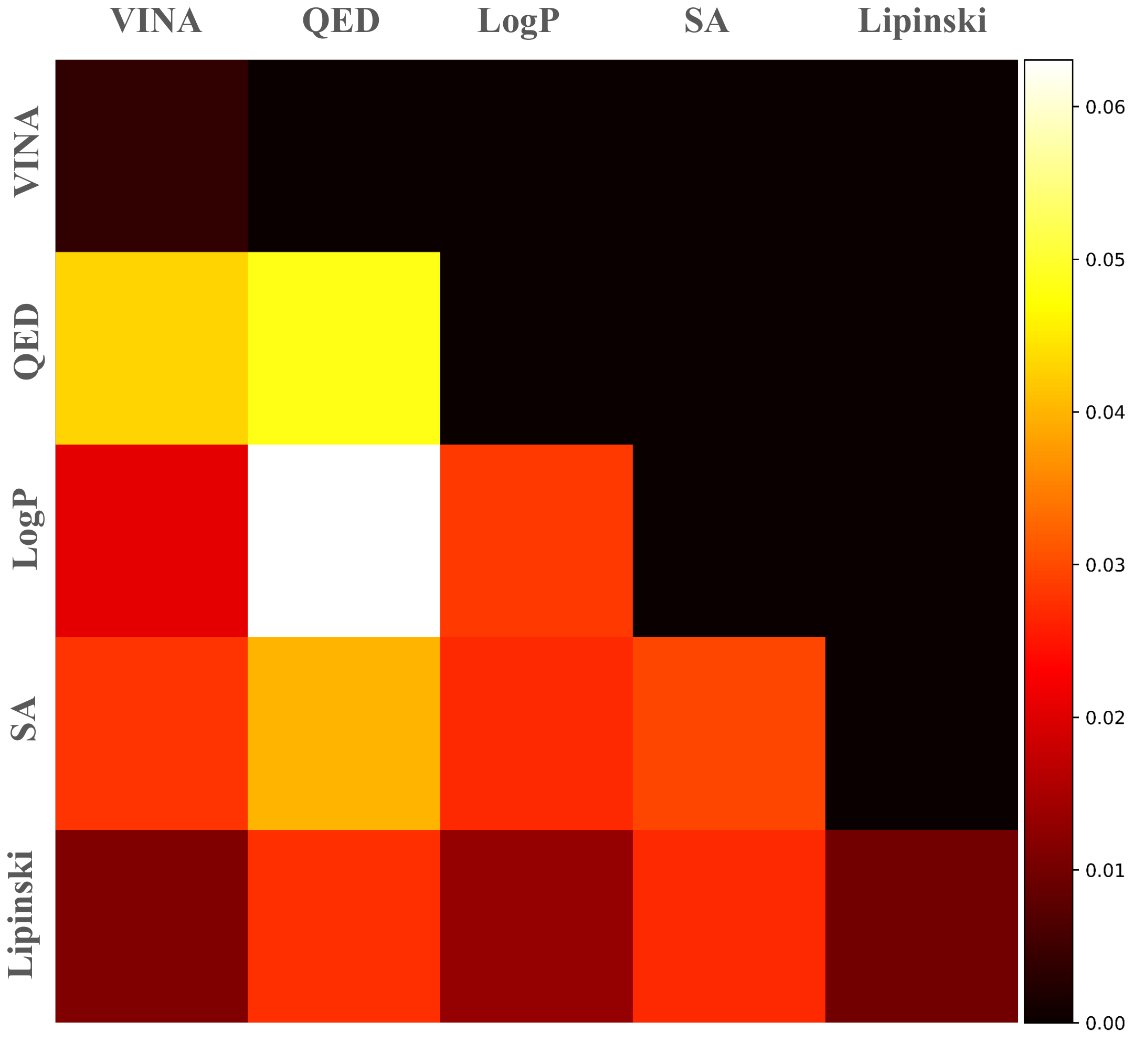}
    \vspace{-3mm}
    \caption{Condition relations.}
    \label{fig:condition_relation}
    \vspace{-3mm}
\end{figure}

\textbf{Results \& Analysis.} 
We visualize the relationship matrix in Figure \ref{fig:condition_relation} to uncover how conditions are interrelated. There are a number of noteworthy observations:(1) The diagonal numbers indicate self-controllability when a single input condition is altered, but the lower left values indicate cross-controllability between many conditions. Cross-control plays a significant role in optimizing molecules properties. (2) The VINA is weakly controlled by itself; instead, it is sensitive to other conditions like QED, LogP, and SA. This discovery explains why changing QED and LogP allows the VINA to exceed SOTA baselines in Table \ref{tab:single_prop} and why the model is poor at controlling individual VINA but works well when controlling multiple properties. A similar explanation holds for the Lipinski improvement achieved by simultaneously changing the QED and SA in Table \ref{tab:multi_prop}. (3) LogP and QED are the most correlated properties. We show the relationships between QED, SA, and LogP in Appendix \ref{ap:correlated_prop}.

\vspace{-4mm}
\section{Conclusion}
\label{sec:conclusions}
\vspace{-2mm}
We propose PrefixMol, a unified model for multi-conditional molecular generation, supporting customized requirements. PrefixMol exhibits good controllability in both single and multi-conditional molecular generation and outperforms previous baselines with the help of controllable generation. More interestingly, we reveal coupling relationships between conditions to provide insights into multi-conditional molecular generation.

\clearpage
\bibliography{references.bib}
\bibliographystyle{icml2023}


\newpage
\appendix
\onecolumn

\section{Related work}
\textbf{Target-aware Molecular Generation.} 
In recent years, various molecular generation methods have attracted extensive attention \citep{gebauer2019symmetry, simm2020generative, simm2020reinforcement, shi2021learning, xu2021end, luo2021predicting, xu2020learning, ganea2021geomol, xu2022geodiff, hoogeboom2022equivariant, jing2022torsional, zhu2022direct, RWtoMGonParagraph1.0, nesterov20203dmolnet,gebauer2022inverse,wu2022diffusion,huang2022mdm,huang20223dlinker,wang2022generative}. However, only some of them could obtain molecules that bind to specific protein targets \citep{imrie2020deep, luo20223d,ragoza2022chemsci,peng2022pocket2mol,liu2022generating}. As shown in Table.\ref{tab: mol_generate_models} (Appendix), we divide target-ware molecular generation models into two types: graph-based and 3D structure-based. Graph-based methods generate molecular graphs given the protein sequence information. For example, SiamFlow \citep{tan2022target} develops a flow model to generate molecular graphs given the targeted protein sequence. To better consider the spatial information, such as spatial isomerism and non-bonded interaction, more 3D structure-based methods have been proposed \citep{imrie2020deep, luo20223d, ragoza2022chemsci, peng2022pocket2mol, liu2022generating}. Among them, Pocket2Mol \citep{peng2022pocket2mol} and GraphBP \citep{liu2022generating} are  representative models to autoregressively generate the atom types and positions, taking the protein pocket as input.

\begin{table}[h]
    \centering
    \caption{Target-aware molecular generation models. }
    \label{tab: mol_generate_models}

        \begin{tabular}{cccccccccc}
            \toprule
            Method                               & Type  & Github                                                        & Year \\ \midrule

            SiamFlow \cite{tan2022target}        & Graph & --                                                            & 2022 \\
            DeLinker \cite{imrie2020deep}        & 3D    & \href{https://github.com/oxpig/DeLinker}{Tensorflow}          & 2020 \\
            Luo's model \cite{luo20223d}         & 3D    & \href{https://github.com/luost26/3D-Generative-SBDD}{PyTorch} & 2021 \\
            LiGAN \cite{ragoza2022chemsci}       & 3D    & \href{https://github.com/mattragoza/LiGAN}{PyTorch}           & 2021 \\
            Pocket2Mol \cite{peng2022pocket2mol} & 3D    & \href{https://github.com/pengxingang/Pocket2Mol}{PyTorch}     & 2022 \\
            GraphBP \cite{liu2022generating}     & 3D    & \href{https://github.com/divelab/GraphBP}{PyTorch}            & 2022 \\
            \bottomrule
        \end{tabular}
\end{table}

\textbf{Target-aware Molecular Generation.} 
In recent years, various molecular generation methods have attracted extensive attention \citep{gebauer2019symmetry, simm2020generative, simm2020reinforcement, shi2021learning, xu2021end, luo2021predicting, xu2020learning, ganea2021geomol, xu2022geodiff, hoogeboom2022equivariant, jing2022torsional, zhu2022direct, RWtoMGonParagraph1.0, nesterov20203dmolnet,gebauer2022inverse,wu2022diffusion,huang2022mdm,huang20223dlinker,wang2022generative}. However, only some of them could obtain molecules that bind to specific protein targets \citep{imrie2020deep, luo20223d,ragoza2022chemsci,peng2022pocket2mol,liu2022generating}. As shown in Table.\ref{tab: mol_generate_models} (Appendix), we divide target-ware molecular generation models into two types: graph-based and 3D structure-based. Graph-based methods generate molecular graphs given the protein sequence information. For example, SiamFlow \citep{tan2022target} develops a flow model to generate molecular graphs given the targeted protein sequence. To better consider the spatial information, such as spatial isomerism and non-bonded interaction, more 3D structure-based methods have been proposed \citep{imrie2020deep, luo20223d, ragoza2022chemsci, peng2022pocket2mol, liu2022generating}. Among them, Pocket2Mol \citep{peng2022pocket2mol} and GraphBP \citep{liu2022generating} are  representative models to autoregressively generate the atom types and positions, taking the protein pocket as input.

\begin{table}[htbp]
    \centering
    \caption{Controllable molecule generation models. }
    \label{tab: mol_controllable_models}

        \begin{tabular}{cccccccccc}
            \toprule
            Method                                  & Type     & Github                                                                                    & Year \\ \midrule

            REINVENT \cite{RWtoMOptParagraph1.2}    & RL       & \href{https://github.com/MarcusOlivecrona/REINVENT}{PyTorch}                              & 2017 \\
            MolDQN \cite{RWtoMOptParagraph1.4}      & RL       & \href{https://github.com/google-research/google-research/tree/master/mol_dqn}{Tensorflow} & 2019 \\
            RationaleRL \cite{RWtoMOptParagraph1.5} & RL       & \href{https://github.com/wengong-jin/multiobj-rationale}{PyTorch}                         & 2020 \\
            CVAE \cite{lim2018molecular}            & SL       & \href{https://github.com/jaechanglim/CVAE}{Tensorflow}                                    & 2018 \\
            CMG \cite{shin2021controlled}           & SL       & \href{https://github.com/deargen/cmg}{Tensorflow}                                         & 2021 \\
            CLaSS \cite{das2021accelerated}        & Opt      & \href{https://github.com/IBM/controlled-peptide-generation}{PyTorch}                       & 2021 \\
            RetMol \cite{wang2022retrieval}         & Retrival & --                                        & 2022 \\
            \bottomrule
        \end{tabular}
\end{table}

\section{GNN layer}
Denote $G$ as a GVP layer and the $l$-th GNN layer, i.e., $\text{GNN}(\mathcal{V}, \vec{\mathcal{V}}, \mathcal{E}, \vec{\mathcal{E}})$, is:

\vspace{-4mm}
\begin{align*}
  \small
  \begin{cases}
      ({{v}}_j', \vec{{v}}_j') = G_{v}({{v}}^{(l-1)}_j, \vec{{v}}^{(l-1)}_j) \\
      ({{e}}_{ij}', \vec{{e}}_{ij}') = G_{e}({{e}}^{(l-1)}_{ij}, \vec{{e}}^{(l-1)}_{ij}) \\
      {{m}}_j' = {{v}}_j' \circ \text{MLP}_1({{e}}_{ij}') \\
      \vec{{m}}_j' = \text{MLP}_2({{e}}_{ij}') \circ \vec{{v}}_{j}' + \text{MLP}_3({{v}}_{j}') \circ \vec{{e}}_{ij}' \\
      ({{m}}^{(l)}_j, \vec{{m}}^{(l)}_j) = G_{m}({{m}}_j', \vec{{m}}_j')\\
      ({{m}}^{(l)}_i, \vec{{m}}^{(l)}_i)=\sum_{j\in {N}_i} G_{m}({{v}}^{(l-1)}_j, \vec{{v}}^{(l-1)}_{j}, {{e}}^{(l-1)}_{ij}, \vec{{e}}^{(l-1)}_{ij}) \\ 
  \end{cases}
\end{align*}

\section{Properties}

{$\bullet$}   \textbf{VINA}: Vina score is a theoretical evaluation of the binding affinity between a small molecule and a target. A molecule with higher affinity is likely to have a higher potential for bioactivity.

    {$\bullet$}   \textbf{QED}: The quantitative estimate of drug-likeness \cite{QED} takes molecular properties into account in order to quantify drug-likeness. It ranges from 0 (all properties unfavorable) to 1 (all properties favorable).

    {$\bullet$}   \textbf{SA}:  Synthetic accessibility score is a measure of the difficulty of synthesizing a chemical, standardized between 0 and 1, with greater values indicating simpler synthesis.

    {$\bullet$}   \textbf{LogP}: LogP, the octanol-water partition coefficient, is a measure of hydrophobicity when one of the solvents is water and the other is a nonpolar solvent. Typically, promising medication candidates should have LogP values \cite{LogP}  between -0.4 and 5.6.

    {$\bullet$}   \textbf{Lipinski}: Lipinski's rule of five is used to determine a drug's similarity to another drug by calculating the number of rules the drug follows \cite{lipinski}.

\clearpage
\section{Further Explanations}

\textbf{The formula of the attention mechanism of the sequence embedding $\textbf{x}$.} \label{ap:prefix_adp}
Here we show the attention mechanism of the sequence embedding $\textbf{x}$ and the attention computation of the prefix features $\mathbf{p}_{\phi}$ in detail.

\begin{equation}
    \begin{aligned}
        \small
        \begin{split}
            & head = \text{Attn}(\mathbf{x}\mathbf{W}_q, \text{Cat}(\mathbf{p}_{\phi} \mathbf{W}_{k}, \mathbf{c}\mathbf{W}_{k}), \text{Cat}(\mathbf{p}_{\phi} \mathbf{W}_{k}, \mathbf{c}\mathbf{W}_v)) \\
            & = \text{softmax}\big(\mathbf{x}\mathbf{W}_q \text{Cat}(\mathbf{p}_{\phi}\mathbf{W}_{k}, \mathbf{c}\mathbf{W}_{k})^\top\big) \begin{bmatrix} \mathbf{p}_{\phi}\mathbf{W}_{v} \\ \mathbf{c}\mathbf{W}_v \end{bmatrix} \\
            & = (1 - \lambda(\mathbf{x})) \text{softmax}(\mathbf{x}\mathbf{W}_q\mathbf{W}_{k}^\top\mathbf{c}^\top)\mathbf{c}\mathbf{W}_v \\& \quad + \lambda(\mathbf{x})\text{softmax}(\mathbf{x}\mathbf{W}_q\mathbf{W}_k^\top\mathbf{p}_{\phi}^\top)\mathbf{p}_{\phi}\mathbf{W}_{v} \\
            & = (1 - \lambda(\mathbf{x})) \underbrace{ \text{Attn}(\mathbf{x}\mathbf{W}_q, \mathbf{c}\mathbf{W}_{k}, \mathbf{c}\mathbf{W}_v) }_{\text{self attention}}\\ & \quad + \lambda(\mathbf{x}) \underbrace{ \text{Attn}(\mathbf{x}\mathbf{W}_q, \mathbf{p}_{\phi} \mathbf{W}_{k}, \mathbf{p}_{\phi} \mathbf{W}_{v}) }_{\text{prefix attention }},
        \end{split}
    \end{aligned}
\end{equation}

\begin{equation}
\begin{aligned}
\small
  \begin{split}
  & head = \text{Attn}(\mathbf{p}_{\phi} \mathbf{W}_q, \text{Cat}(\mathbf{p}_{\phi} \mathbf{W}_k, \mathbf{c}\mathbf{W}_{k}), \text{Cat}(\mathbf{p}_{\phi} \mathbf{W}_v, \mathbf{c}\mathbf{W}_v)) \\
  & = (1 - \lambda(\mathbf{p}_{\phi})) \text{Attn}(\mathbf{p}_{\phi} \mathbf{W}_q, \mathbf{c}\mathbf{W}_{k}, \mathbf{c}\mathbf{W}_v) \\ & \quad + \lambda(\mathbf{p}_{\phi})  \text{Attn}(\mathbf{p}_{\phi} \mathbf{W}_q, \mathbf{p}_{\phi}\mathbf{W}_k, \mathbf{p}_{\phi}\mathbf{W}_v) \\
  & = \underbrace{\text{Attn}(\mathbf{p}_{\phi}\mathbf{W}_q, \mathbf{p}_{\phi}\mathbf{W}_k, \mathbf{p}_{\phi}\mathbf{W}_v)}_{\text{prefix correlation}}
  \end{split}
\end{aligned}
\end{equation}

Where $\text{Cat}$ is the concatenate operation, and $\lambda(\mathbf{x})$ shown below is a scalar that represents the sum of normalized attention weights on the prefixes.

\begin{equation}
\begin{aligned}
  \small
  \lambda(\mathbf{x}) = \frac{\sum_i\exp (\mathbf{x}\mathbf{W}_q \mathbf{W}_{k}^\top\mathbf{p}_{\phi}^\top)_i}{\sum_i \exp (\mathbf{x}\mathbf{W}_q \mathbf{W}_{k}^\top\mathbf{p}_{\phi}^\top)_i + \sum_j \exp(\mathbf{x}\mathbf{W}_q\mathbf{W}_{k}^\top\mathbf{c}^\top)_j}.
\end{aligned}
\end{equation}

\clearpage
\section{Additional Results}
\subsection{Molecule Design} \label{ap:molecule_design}

\begin{figure*}[htbp]
    \centering
    \includegraphics[width=0.84\textwidth]{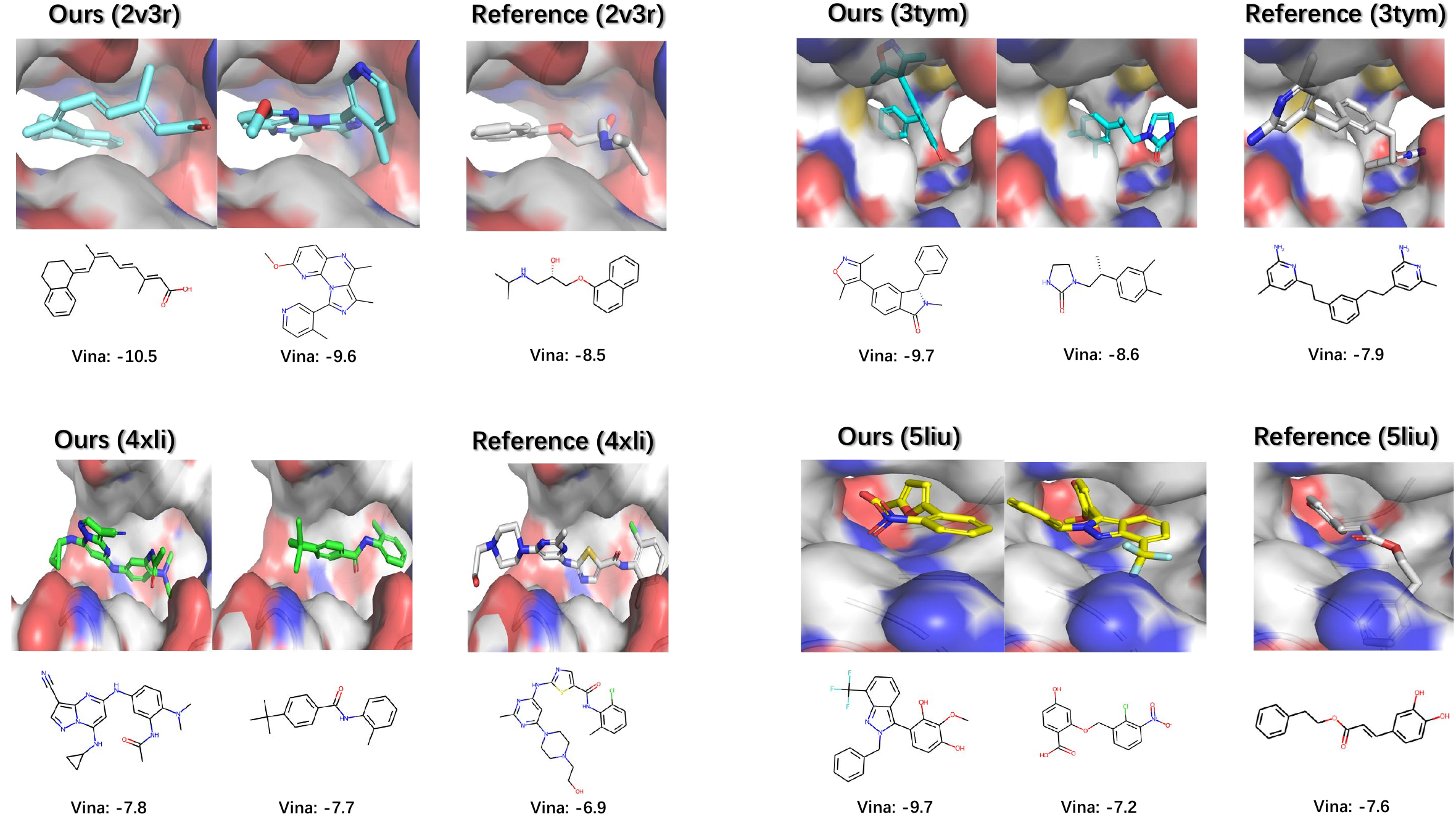}
    \caption{Examples of 3D-generated compounds with better binding affinity than reference molecules are depicted. The lower the Vina score, the greater the binding affinity.}
    \label{tab:cases}
\end{figure*}

We present examples of generated molecules by our method as four case studies shown in Figure.\ref{tab:cases}.

\subsection{Correlated Properties} \label{ap:correlated_prop}
\begin{figure*}[b]
    \centering
    \includegraphics[width=0.84\textwidth]{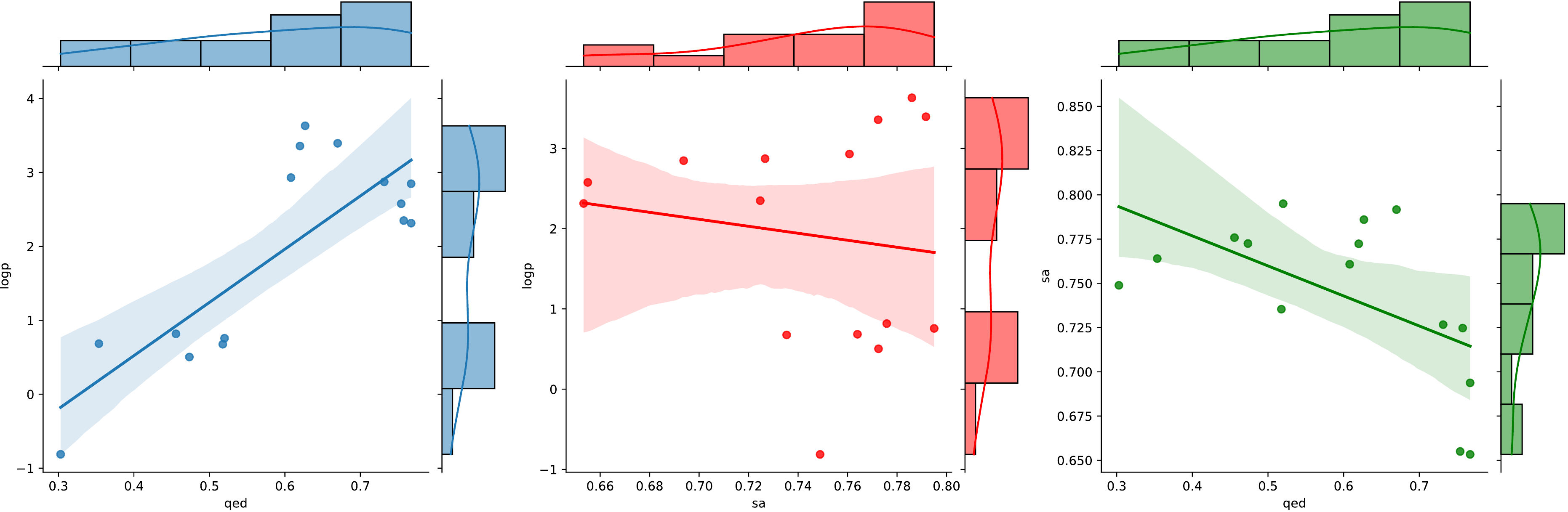}
    \caption{Correlation between QED, SA, and LogP.}
    \label{tab:qed_logp_sa}
\end{figure*}

As Table.\ref{tab:single_prop} shows, QED, SA, and LogP follow the conditional inputs well. Therefore, we reveal them in Figure.\ref{tab:qed_logp_sa} to further explore their potential relationship. We can clearly see that the slope of the regression curve of QED and LogP is high and positively correlated.  

\end{document}